\documentclass[letterpaper, 10 pt, conference]{ieeeconf}  

\IEEEoverridecommandlockouts                              

\usepackage[linesnumbered,ruled,vlined]{algorithm2e}
\usepackage{graphics} 
\usepackage{epsfig} 
\usepackage{mathptmx} 
\usepackage{times} 
\usepackage{amsmath} 
\usepackage{amssymb}  
\usepackage{subfigure}
\usepackage{epstopdf}
\usepackage{float}
\usepackage{bm}
\graphicspath{{./fig/}}
\usepackage{color}
\usepackage{multirow}
\usepackage[hidelinks]{hyperref}

\usepackage{xifthen}
\usepackage{xparse}
\usepackage{subdepth}
\usepackage{leftidx}
\usepackage{bm}
\usepackage{amsmath}

\newcommand{\bbm}{\begin{bmatrix}}
	\newcommand{\ebm}{\end{bmatrix}}

\DeclareMathAlphabet{\mbf}{OT1}{ptm}{b}{n}
\newcommand{\mbs}[1]{{\bm{#1}}}

\newcommand{\mbsbar}[1]{{\overline{\boldsymbol{#1}}}}
\newcommand{\mbshat}[1]{{\hat{\boldsymbol{#1}}}}
\newcommand{\mbstilde}[1]{{\tilde{\boldsymbol{#1}}}}

\newcommand{\mbsdot}[1]{{\dot {\boldsymbol{#1}}}}

\newcommand{\mbfbar}[1]{{\overline{\mbf{#1}}}}
\newcommand{\mbfhat}[1]{{\hat{\mbf{#1}}}}
\newcommand{\mbftilde}[1]{{\tilde{\mbf{#1}}}}

\newcommand{\mbfdot}[1]{{\dot{\mbf{#1}}}}

\newcommand{\cframe}[1]{{\smash{\protect\underrightarrow{\mathcal{F}}_{#1}}}}

\DeclareMathAlphabet{\mathbfit}{OML}{cmm}{b}{it}
\newcommand{\homo}[1]{{\mathbfit{#1}}}

\newcommand{\mbfh}[1]{{\homo{#1}}}

\newcommand{\trans}[3]{\leftidx{_{#1}}{\mbf r}{\IfValueTF{#2}{_{#2#3\hspace{2pt}}}{}}} 

\newcommand{\vel}[3]{\leftidx{_{#1}}{\mbf v}{\IfValueTF{#2}{_{#2#3\hspace{2pt}}}{}}} 
\newcommand{\veltilde}[3]{\leftidx{_{#1}}{\mbftilde v}{\IfValueTF{#2}{_{#2#3\hspace{2pt}}}{}}} 
\newcommand{\velbar}[3]{\leftidx{_{#1}}{\mbfbar v}{\IfValueTF{#2}{_{#2#3\hspace{2pt}}}{}}} 
\newcommand{\velhat}[3]{\leftidx{_{#1}}{\mbfhat v}{\IfValueTF{#2}{_{#2#3\hspace{2pt}}}{}}} 
\newcommand{\veldot}[3]{\leftidx{_{#1}}{\mbfdot v}{\IfValueTF{#2}{_{#2#3\hspace{2pt}}}{}}} 

\newcommand{\acc}[3]{\leftidx{_{#1}}{\mbf a}{\IfValueTF{#2}{_{#2#3\hspace{2pt}}}{}}} 
\newcommand{\acctilde}[3]{\leftidx{_{#1}}{\mbftilde a}{\IfValueTF{#2}{_{#2#3\hspace{2pt}}}{}}} 
\newcommand{\accbar}[3]{\leftidx{_{#1}}{\mbfbar a}{\IfValueTF{#2}{_{#2#3\hspace{2pt}}}{}}} 

\newcommand{\rotvel}[3]{\leftidx{_{#1}}{\mbs \omega}{\IfValueTF{#2}{_{#2#3\hspace{2pt}}}{}}} 
\newcommand{\rotveltilde}[3]{\leftidx{_{#1}}{\mbstilde \omega}{\IfValueTF{#2}{_{#2#3\hspace{2pt}}}{}}} 
\newcommand{\rotvelbar}[3]{\leftidx{_{#1}}{\mbsbar \omega}{\IfValueTF{#2}{_{#2#3\hspace{2pt}}}{}}} 
\newcommand{\rotvelhat}[3]{\leftidx{_{#1}}{\mbshat \omega}{\IfValueTF{#2}{_{#2#3\hspace{2pt}}}{}}} 
\newcommand{\rotveldot}[3]{\leftidx{_{#1}}{\mbsdot \omega}{\IfValueTF{#2}{_{#2#3\hspace{2pt}}}{}}} 

\newcommand{\C}[2]{\leftidx{}{\mbf C}{_{#1#2\hspace{2pt}}}} 
\newcommand{\T}[2]{\leftidx{}{\mbfh T}{_{#1#2\hspace{2pt}}}} 

\newcommand{\generalTwo}[2]{{\mbf #1}{_{#2}}} 
\newcommand{\generalThree}[3]{\leftidx{_{#1}}{\mbf #2}{_{#3}}} 
\newcommand{\pixel}[1]{{\mbfh u}{_{#1}}} 

\def \onecolume{8.5cm}

\title{\LARGE \bf
MID-Fusion: Octree-based Object-Level Multi-Instance Dynamic SLAM
}

\author{Binbin Xu, Wenbin Li, Dimos Tzoumanikas,  Michael Bloesch, Andrew Davison, Stefan Leutenegger 
\thanks{The authors are with Department of Computing, Imperial College London, United Kingdom.
 	{\tt\small \{b.xu17, wenbin.li, dt214, m.bloesch, a.davison, s.leutenegger\}@imperial.ac.uk}}
 	\thanks{Video link: \url{https://youtu.be/gturboNl9gg}}
 	}

\begin{document}
\maketitle
\thispagestyle{empty}
\pagestyle{empty}


\begin{abstract}
We propose a new multi-instance dynamic RGB-D SLAM system using an object-level octree-based volumetric representation. It can provide robust camera tracking in dynamic environments and at the same time, continuously estimate geometric, semantic, and motion properties for arbitrary objects in the scene. For each incoming frame, we perform instance segmentation to detect objects and refine mask boundaries using geometric and motion information. Meanwhile, we estimate the pose of each existing moving object using an object-oriented tracking method and robustly track the camera pose against the static scene. Based on the estimated camera pose and object poses, we associate segmented masks with existing models and incrementally fuse corresponding colour, depth, semantic, and foreground object probabilities into each object model. In contrast to existing approaches, our system is the first system to generate an object-level dynamic volumetric map from a single RGB-D camera, which can be used directly for robotic tasks. Our method can run at 2-3 Hz on a CPU, excluding the instance segmentation part. We demonstrate its effectiveness by quantitatively and qualitatively testing it on both synthetic and real-world sequences.

\end{abstract}

\section{Introduction}
In Simultaneous Localisation and Mapping (SLAM) both, the map of the unknown environment as well as the robot pose within it, are concurrently estimated from the data of its on-board sensors only. In recent years, the field of SLAM has experienced rapid progress. It started from sparse SLAM~\cite{Davison:etal:PAMI2007, Klein:Murray:ISMAR2007}, and evolved into dense SLAM~\cite{Newcombe:etal:ISMAR2011} thanks to the increased computational power of GPU and affordability of depth sensors. More recently, many people have begun to leverage Deep Neural Networks and their ability to learn from large amounts of training data to improve SLAM. This fast-evolving research in SLAM has, since then, lead to strong progress in various fields of applications, such as robotics, Virtual Reality (VR), and Augmented Reality (AR). 

Despite this progress, much work is still based on the fundamental assumption of a static environment, within which points in the 3D world always maintain the same spatial position in the global world, with the only moving object being the camera. This assumption enabled the success of early phases of development as it alleviated the chicken-and-egg problem between map estimation and sensor pose estimation. A camera pose can be estimated between a live frame and a reference frame, which is based on the assumption that the relative transformation between those two images is caused only by the camera motion. It is this basic, yet strong, assumption that allowed a joint probabilistic inference (sparse SLAM~\cite{Durrant-Whyte:etal:RAM2006}) or an alternating optimisation (dense SLAM~\cite{Engel:etal:PAMI2017}) of map and pose relationship to solve SLAM. Any moving objects in the environment would be treated as outliers to the static model and are intentionally ignored by tracking and mapping. 

This idealised setup, therefore, can only handle a small amount of dynamic elements and disqualifies itself from many real-world applications as environments, especially where humans are present, change constantly. A robust SLAM system, which works in highly dynamic environments, is still an open problem, which we seek to address in this work. 

\begin{figure}[tbp]
	\centering
	\includegraphics[width=\linewidth]{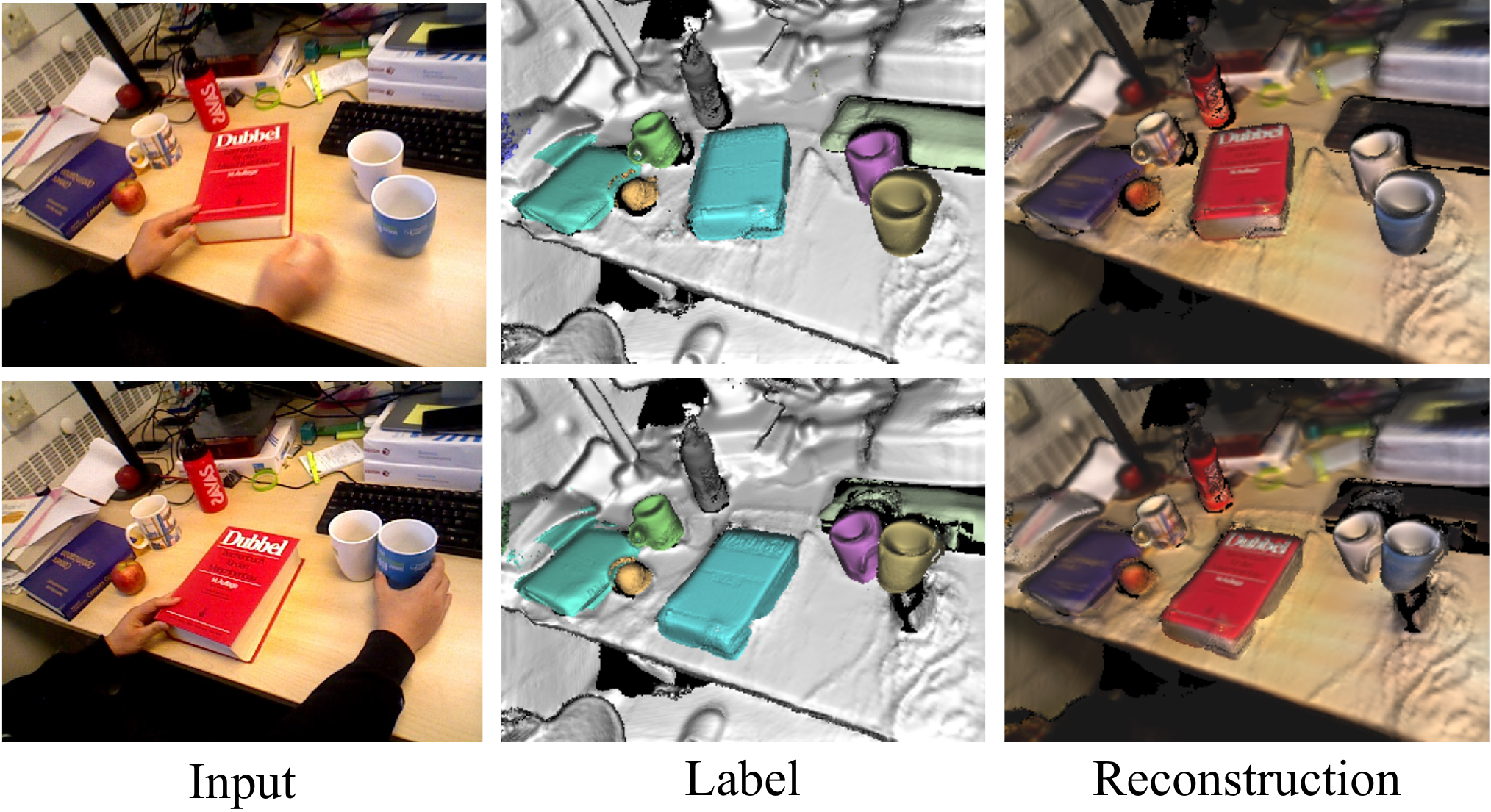}
	\vspace{-7mm}
	\caption{An overview of our system. Given RGB-D images, our system builds an object-level dense volumetric map that deals with dynamic objects and ignores people. Next to the input image we show the labelled object models as well as the coloured reconstruction.}
	\label{fig:overview}
	\vspace{-4mm}
\end{figure} 

Although dynamic SLAM has been studied for a couple of decades~\cite{Wang:etal:ICRA2003}, approaches based on visual dense SLAM have only recently been explored. They can be categorised into three main directions. One deforms the whole world in a non-rigid manner in order to include a deformable/moving object~\cite{Newcombe:etal:CVPR2015}. The second specifically aims at building a single static background model, while ignoring all possibly moving objects and thus improving the accuracy of camera tracking~\cite{Jaimez:etal:ICRA2017, Scona:etal:ICRA2018,Barnes:etal:ICRA2018,Bescos:etal:RAL2018}. The third models dynamic components by creating sub-maps for every possibly rigidly moving object in the scene while fusing corresponding information into these sub-maps~\cite{Runz::Agapito::ICRA2017, Barsan:etal:ICRA2018,  Runz::Agapito::ARXIV2018}. We are more interested in the third direction since we believe that, similar to human perception, an awareness of instances in the map would be a more proper solution for robots to perceive the changing environment and has higher potential to achieve a meaningful map representation. However, most existing approaches build maps using a collection of surfels, which is difficult to be used directly for robotic tasks. The only two systems that support sub-map volumetric map, we know of so far, are \cite{Barsan:etal:ICRA2018} and \cite{McCormac:etal:3DV2018}. However, the former has been specifically designed for an outdoor stereo camera setting and the latter only deals with static environments. Here, we propose the first object-level dynamic volumetric map for indoor environment applications, where free space and surface connectivity can be represented for each object model. We further improve its memory efficiency by utilising an octree-based structure. Despite showing some promising results based on deep learning, most methods \cite{Runz::Agapito::ICRA2017, Runz::Agapito::ARXIV2018, Barsan:etal:ICRA2018} simply leverage predictions from neural network without much refinement in the map fusion. In this paper, we integrate and refine semantic predictions by fusing them into object models.

The main contributions in this paper are divided into four main parts. We propose 
\begin{enumerate}
	\item the first RGB-D multi-instance dynamic SLAM system using a volumetric representation,
	\item a more robust tracking method utilising weighting via measurement uncertainty and being re-parametrised for object tracking,
	\item an integrated segmentation using geometric, photometric, and semantic information,
	\item a probabilistic fusion of semantic distribution and a foreground object probability into octree-based object models.
\end{enumerate}

%
\section{Related work}
In the majority of SLAM systems the environment is assumed to be static. To tackle dynamic environment in real-world applications, several solutions have recently been proposed and they can be mainly categorised into three directions as introduced in last section. We will introduce and compare the last two types of approaches in further details in this section. One straightforward way for dynamic SLAM is to segment dynamic objects out as outliers and intentionally ignore them from tracking and reconstruction to avoid corruption in the pose estimation. StaticFusion~\cite{Scona:etal:ICRA2018} performs segmentation by coupling camera motion residuals, depth inconsistency and a regularisation term. Barnes \textit{et al.}~\cite{Barnes:etal:ICRA2018} learn to segment possibly moving objects in a self-supervised way, which is limited by the availability of training data and may often misclassify static objects. Bescos \textit{et al.}~\cite{Bescos:etal:RAL2018} combine Mask-RCNN~\cite{He:etal:ICCV2017} with depth inconsistency checking to segment moving objects and further inpaint those areas with static background. Those methods provide a more robust approach in dynamic scene than conventional SLAM methods, however, information regarding the moving objects is lost. Instead, our approach aims to simultaneously track and reconstruct static background and dynamic and static objects in the scene, while at the same time, provide state-of-the-art tracking accuracy.

There are three approaches, to our knowledge, which provide similar functionality as ours and can reconstruct multiple moving objects in the scene -- the third way to tackle dynamic SLAM. Co-Fusion~\cite{Runz::Agapito::ICRA2017} segments objects by either ICP motion segmentation or semantic segmentation and then tracks objects separately based on ElasticFusion~\cite{Whelan:etal:IJRR2016}. MaskFusion~\cite{Runz::Agapito::ARXIV2018} segments objects using a combination of instance segmentation from Mask-RCNN and geometric edges, and tracks objects using the same approach as Co-Fusion. Both Co-Fusion and MaskFusion use surfels to represent map models, which is memory efficient but cannot directly provide free space information in the map, and neither surface connectivity. DynSLAM~\cite{Barsan:etal:ICRA2018} focuses on outdoor environments using stereo cameras. In contrast, our system focuses on indoor environments consisting of many (potentially) moving objects using a single RGB-D camera. 

In terms of differences in system components, our system further differentiates itself from above approaches. In camera tracking, we weighted photometric and geometric terms by their measurement uncertainty, instead of a single weight such as in~\cite{Whelan:etal:IJRR2016}. Also, to be robust to depth loss, we derive two terms from different frames to complement one another.
To track objects, all previous methods use a virtual camera pose, which is not robust to object rotation due to its difficulty to converge. We found best robustness by re-parametrising it into object coordinate. To generate object masks, we combine both information to provide better boundary conditions, instead of using just motion or just semantic information. When fusing information to object models, we fuse not only depth and colour information, but also semantic and foreground predictions while previous methods just take predictions from neural network without any refinement. In terms of speed, all above three requires one even two powerful GPUs, while our method, despite running only on CPU, is capable of performing a similar speed to DynSLAM~\cite{Barsan:etal:ICRA2018}. 

Another very recent work related to ours is Fusion++\cite{McCormac:etal:3DV2018}, which generates an object-level volumetric map yet in static environments. In addition to handling dynamic scenes, our system utilises a joint photometric and geometric tracking to robustly track both camera and object poses while Fusion++ only use geometric tracking to estimate camera pose. Furthermore, to have a better object mask boundary for fusion and tracking, we combine geometric, motion and existing model information to refine mask boundary instead of directly using predicted mask as was done in Fusion++. In terms of map representation, Fusion++ is based on discrete voxel grids, which suffers from scalability issues, while we represent all our object models in memory-efficient octree structures. 

\section{Notations and preliminaries}
In this paper, we will use the following notation: a reference coordinate frame is denoted $\cframe{A}$.  The homogeneous transformation from $\cframe{B}$ to $\cframe{A}$ is denoted as $\T{A}{B}$, which is composed of a rotation matrix $\C{A}{B}$ and a translation vector $\trans{A}{A}{B}$. For each pair of images, we distinguish them as live ($L$) and reference ($R$) image. For example, a live RGB-D image contains the intensity image $I_L$ and depth image $D_L$, with 2D pixel positions denoted as $\pixel{L}$ and pixel lookup (including bilinear interpolation) denoted as $[\cdot]$. Perspective projection and back-projection are denoted $\pi$ and $\pi^{-1}$, respectively.

In our system, we store every detected object into a separate object coordinate frame $\cframe{\generalTwo{O}{n}}$, with $n \in \{ 0\ldots, N\}$ where $N$ is the total number of objects (excluding background) and 0 denotes background. We assume a canonical static volumetric model is stored in each object coordinate frame, forming the basis of our multi-instance SLAM system. In addition, each object is also associated with a COCO dataset~\cite{Lin:etal:ECCV2014} semantic class label $c_n \in \{0, \ldots, 80\}$, a probability distribution over its potential semantic class labels, a current pose w.r.t.\ the world coordinate $\T{W}{O_n}$, and a binary label $s \in \{0, 1\}$ denoting whether the object is believed to be in motion or not. Each object is represented in an separate octree structure, where every voxel stores Signed Distance Function (SDF) value, intensity, foreground probability and the corresponding weights.

\section{Method}
\subsection{System overview}
\begin{figure}[tbp]
	\centering
	\includegraphics[width=\linewidth]{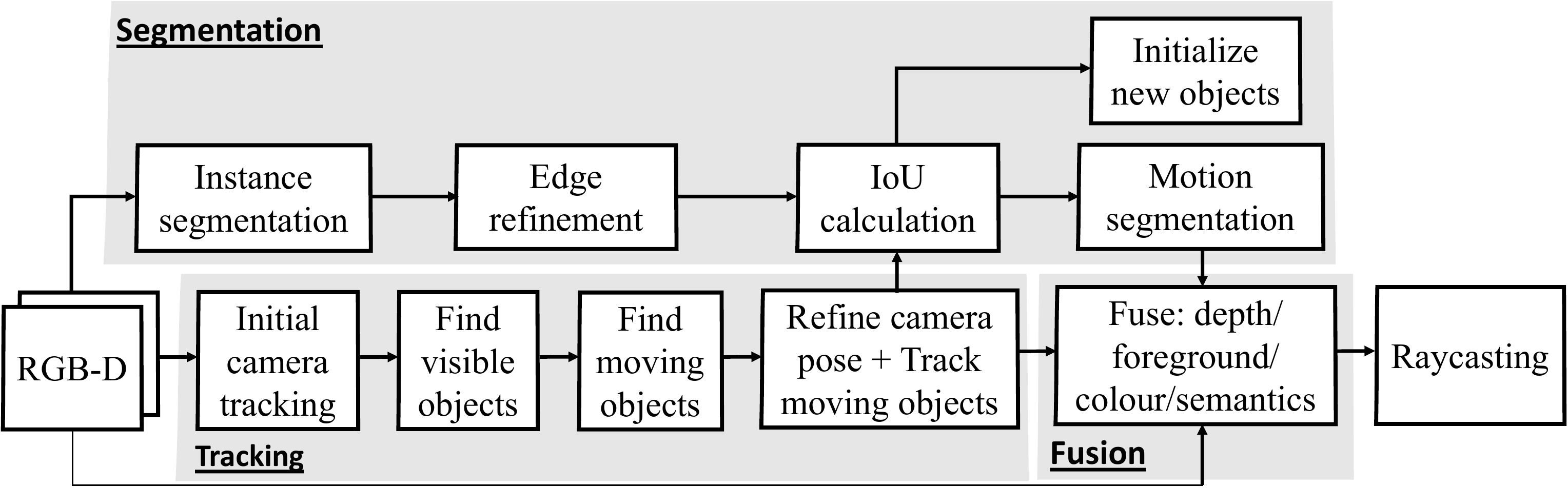}
	\vspace{-6mm}
	\caption{The pipeline of the proposed method}
	\vspace{-5mm}
	\label{fig:pipeline}
\end{figure}

Fig.~\ref{fig:pipeline} shows the pipeline of our proposed system. It is composed of four parts: segmentation, tracking, fusion and raycasting. Each input RGB-D image is processed by Mask R-CNN to perform instance segmentation, which is followed by geometric edge segmentation and motion residuals from tracking to refine mask boundaries (Section~\ref{section: Combined semantic-geometric-motion segmentation}). For the tracking, we first track the camera against all vertices excluding the human mask area (Section~\ref{section: RGB-D Camera tracking}) and then raycast from this pose to find which objects are currently visible in this frame. This can also help associating local object masks with existing object models. We evaluate motion residuals for each object to determine if it is in motion or not, then track moving objects (Section~\ref{section: Object pose estimation}) and refine the camera pose against the static world -- which includes currently static objects (Section~\ref{section: RGB-D Camera tracking}). Using estimated poses of the camera and objects, depth and colour information, as well as predicted semantic and foreground probabilities are fused into the object models (Section~\ref{section: Object-level fusion}). Detection of visible objects as well as raycasting is explained in Section~\ref{section: Raycasting}.

\subsection{RGB-D Camera tracking}
\label{section: RGB-D Camera tracking}
This part estimates the live camera pose $\T{W}{C_L}$ and is composed of two steps. First, it tracks against all model vertices while masking out detected people; second, it tracks against all static scene parts. Both steps are conducted by minimising the dense point-to-plane ICP residual $e_{\mathrm{g}}$ and photometric (RGB) residual $e_{\mathrm{p}}$, which are weighted by individual measurement uncertainty, $w_{\mathrm{g}}$ and $w_{\mathrm{p}}$.
\begin{align}
\label{eq: joint-residual}
E_{\mathrm{track}}(\T{W}{C_L}) &= \frac{1}{2} \left( \sum_{\pixel{L} \in M_L} w_{\mathrm{g}} \, \rho (e_{\mathrm{g}})+ \sum_{\pixel{R} \in M_R} w_{\mathrm{p}} \, \rho (e_{\mathrm{p}}) \right), 
\end{align}
where $\rho$ represents the Cauchy loss function and $M$ is a mask excluding invalid correspondences (for ICP), occlusions (for RGB), and humans. 

For the ICP residual, we use the method proposed in~\cite{Newcombe:etal:ISMAR2011} to minimise point-plane depth error between the live depth map and the rendered depth map of the model on the reference frame:
\begin{align}
\label{eq: icp-residual}
e_{\mathrm{g}}(\T{W}{C_L}) &= \generalThree{W}{n}{}^r [\pixel{R}] \cdot ( \T{W}{C_L} \generalThree{C_L}{v}{} [\pixel{L}] - \generalThree{W}{v}{}^r [\pixel{R}]),
\end{align}
where $\generalThree{C_L}{v}{}$ is live vertex map in the camera coordinate by back-projection and $\generalThree{W}{v}{}^r$ and $\generalThree{W}{n}{}^r$  are the rendered vertex map and normal map expressed in world coordinates. For each pixel $\pixel{L}$ on the live depth map, its correspondence $\pixel{R}$ on the rendered depth map can be found using projective data association:
\begin{align}
\pixel{R} &= \pi(\T{W}{C_R}^{-1} \T{W}{C_L}(\pi^{-1}(\pixel{L}, D_L[\pixel{L}]))),
\end{align}
where $\T{W}{C_R}$ is the camera pose of the reference frame.

For maximum robustness, we combine the ICP residual with a photometric one by rendering a depth map from model in the reference frame and using that depth map to align photometric consistency:
\begin{align}
e_{\mathrm{p}}(\T{W}{C_L})
= I_R[\pixel{R}] - I_L[\pi(\T{W}{C_L}^{-1}( \T{W}{C_R}\pi^{-1}(\pixel{R}, D_R^r[\pixel{R}])))].
\label{eq:rgb-residual}
\end{align}
Different from previous approaches~\cite{Runz::Agapito::ICRA2017}, we evaluate the photometric residuals using rendered reference depth map other than the raw depth map on live frame or reference frame for the de-noised depth quality from models. This choice furthermore improves the robustness of tracking when raw input depth is not available, e.g.\ when the camera is too close to a surface. 

We further introduce a measurement uncertainty weight to combine ICP and RGB residuals. For RGB residuals, the measurement uncertainty is assumed to be constant for all pixels. For ICP residuals, the quality of input depth map is related to the structure of the depth sensor and the depth range. We adopted the inverse covariance definition for depth measurement uncertainty in~\cite{Laidlow::etal::IROS2017}. Given the sensor parameters, i.e.\ baseline $b$, disparity $d$, focal length $f$, and the uncertainties in the x-y plane $\sigma_{xy}$ and disparity direction $\sigma_{z}$, the standard deviation $\sigma_{D}$ for depth sensor measurement in the x, y, z coordinates can be modelled as:
\begin{align}
\label{standard-deviation-for-icp}
\sigma_{D} = (\frac{D_L[\pixel{L}]}{f}\sigma_{xy}, \frac{D_L[\pixel{L}]}{f}\sigma_{xy}, \frac{D_L^2[\pixel{L}]}{fb}\sigma_{z}).
\end{align}
The weight for ICP residuals using the inverse covariance of measurement uncertainty is then defined as:
\begin{align}
w_{\mathrm{g}} = \frac{1}{(\generalThree{W}{n}{}^r)^T\generalThree{W}{n}{}^r\sigma_{D}^T\sigma_{D}}.
\end{align}

The cost function is minimised using the Gauss-Newton approach in a three-level coarse-to-fine scheme. The necessary Jacobians are omitted for space constraint.

After performing an initial camera tracking, we raycast to find visible objects in the view. To find which objects are in motion, we evaluate $E_{\mathrm{track}}(\T{W}{C_L})$ once again on the finest level on the \emph{live} frame. To this end, the RGB residual needs to be re-formulated as:
\begin{align}
e_{\mathrm{p}}(\T{W}{C_L})
= I_L[\pixel{L}] - I_R[\pi(\T{W}{C_R}^{-1}( \T{W}{C_L}\pi^{-1}(\pixel{L}, D_L^r[\pixel{L}])))].
\label{eq:rgb-motion-residual}
\end{align}
We apply a threshold to the combined residual $E_{\mathrm{track}}(\T{W}{C_L})$ to find the motion inliers. If the inlier ratio is lower than 0.9 in the object's rendered mask, then we consider that object is moving and refine its pose as described in Section~\ref{section: Object pose estimation}. The camera pose is then refined by tracking against only static objects using the same objective function and optimisation strategy explained above.

\subsection{Object pose estimation}
\label{section: Object pose estimation}


In this part, we describe how to estimate the pose of moving objects. As opposed to virtual camera based tracking~\cite{Runz::Agapito::ICRA2017,Barnes:etal:ICRA2018}, we propose to employ an object-centric approach, which is less prone to bad initial pose guesses.
We still use a joint dense ICP and RGB tracking, weighted in the same way as Eq.~\ref{eq: joint-residual}, just with different ICP and RGB residual definitions. In the present formulation, we estimate the current relative pose between object and camera, $\T{C_L}{O_L}$, by aligning the live vertex map expressed in the live object frame with the rendered vertex map expressed in the reference object frame:
\begin{align}
\label{eq: object-icp-residual}
e_{\mathrm{g}}(\T{C_L}{O_L}) &= \C{W}{O_R}^{-1} \generalThree{W}{n}{}^r [\pixel{R}] \cdot ( \T{C_L}{O_L}^{-1} \generalThree{C_L}{v}{} [\pixel{L}] - \T{W}{O_R}^{-1} \generalThree{W}{v}{}^r [\pixel{R}]).
\end{align}
The formulation is based on the assumption that each object coordinate frame yields a static canonical object model and thus the point clouds must align. 
The proposed parameterisation leads to more stable tracking due to a smaller lever arm effect of the rotation. When computing the partial derivative of the above cost w.r.t. the rotation \cite{Bloesch:etal:CoRR2016} we get a term proportional to $\C{C_L}{O_L}^{-1} (\generalThree{C_L}{v}{} [\pixel{L}] -\trans{C_L}{C_L}{O_L})$ which is small since we choose the object frame to be centred.
In analogy, we also re-formulate the RGB residual as:
\begin{align}
e_{\mathrm{p}}(\T{C_L}{O_L}) 
= I_R[\pixel{R}] - I_L[\pi(\T{C_L}{O_L} \T{C_R}{O_R}^{-1}(\pi^{-1}(\pixel{R}, D_R^r[\pixel{R}])))].
\label{eq:object-rgb-residual}
\end{align}
The above cost function is also optimised using Gauss-Newton approach in a three-level coarse-to-fine scheme with $\T{C_L}{O_L}$ initialised as $\T{C_L}{O_R}$.

\subsection{Combined semantic-geometric-motion segmentation }
\label{section: Combined semantic-geometric-motion segmentation}
For each RGB-D frame, we use Mask R-CNN~\cite{He:etal:ICCV2017} to find semantic instances, followed by geometric edge refinement to solve leaked mask boundary~\cite{Runz::Agapito::ARXIV2018}. Then we render instance masks for each map object to the live frame by means of raycasting. We associate local segmentation masks, which are generated from Mask R-CNN and geometric refinement, with existing object models by calculating the intersection of union (IoU) with the rendered masks. We assign the local segmentation mask to the rendered mask which has the largest intersection and where the intersection is larger than 0.5. In comparison to~\cite{Runz::Agapito::ARXIV2018}, we do not require predicted semantic label of the local segmentation mask to be the same as object semantic class since the prediction may be subject to high uncertainty. Instead, we trust probabilistic fusion of semantic predictions to refine the objects' semantic labels (described in Section~\ref{section: Object-level fusion}).

For segmentation masks that do not belong to any existing objects, a new object model will be initialised (described in Section~\ref{section: Object-level fusion}). For objects without associated local segmentation masks, i.e. Mask R-CNN has no corresponding detection, we choose its rendered mask from the model for the subsequent fusion process.

After associating segmentation masks with object models, we further refine the segmentation masks based on motion residuals of object tracking. We evaluate Eq.~(\ref{eq: joint-residual}) again on the finest level, however, this time we evaluate photometric residual on the live frame:
\begin{align}
e_{\mathrm{p}}(\T{W}{C_L}) 
= I_L[\pixel{L}] - I_R[\pi(\T{C_R}{O_R} \T{C_L}{O_L}^{-1}(\pi^{-1}(\pixel{L}, D_L^r[\pixel{L}])))].
\label{eq:object-rgb-motion-residual}
\end{align}
Pixels whose joint ICP and RGB residuals are too high are treated as outliers and filtered out in the segmentation mask.


Before integration, we also generate a foreground mask based on the local segmentation mask. The use of foreground probabilities is inspired by the foreground/background probabilities introduced in~\cite{McCormac:etal:3DV2018} and allows to avoid spurious integration due to wrong segmentation masks. Information in both foreground and background regions are integrated into the models. In order to avoid impairing the efficiency of the octree structure, we use dilated segmentation masks as background mask. Pixels in the foreground are assigned an foreground probability of 1.0 while pixels in the dilated background are assigned 0. For undetected existing objects that Mask R-CNN fails on, we assign an foreground probability of 0.5 to their foreground due to their lower possibility of existence. 

\subsection{Object-level fusion}
\label{section: Object-level fusion}
From each frame, we integrate depth, colour, semantics and foreground probability information into object models using foreground and background masks.
Using the relative pose $\T{O_n}{C_L}$ and depth, the Truncated SDF (TSDF) is updated following the approach of Vespa et al.~\cite{Vespa:etal:RAL2018}. Concurrently within the same voxels, colour and foreground probability are updated using a weighted average. For semantic fusion, we refine the semantic class probability distribution for each model using averaging, instead of Bayesian updating which often leads to overconfidence when used with Mask R-CNN predictions\cite{McCormac:etal:3DV2018}. 

%

For every segmentation mask that cannot be associated with any existing objects, we initialise a new object model whose coordinate frame is centred around the object itself. We back-project all points in the mask into world coordinates and then find the centre and size of these point clouds. To account for possible occlusions, we initialise the TSDF volume size to be 3 times the point cloud size to avoid additional padding. We choose the volume resolution such that each voxel size is slightly bigger than 1mm in order to support detailed object reconstruction. With the octree-based structure, the unused voxels will not be initialised and the whole system remains memory-efficient. The initial object translation in $\T{W}{O_n}$ is chosen as the left side corner of the object volume and the orientation is aligned with world coordinates.

\subsection{Raycasting}
\label{section: Raycasting}
For raycasting, we use a similar method as proposed in~\cite{McCormac:etal:3DV2018}. However, as shown in the pipeline Fig.~\ref{fig:pipeline}, our system involves at least four raycasting operations: depth rendering in tracking, finding visible objects, IoU calculation, and visualisation, which will be computationally expensive if we continuously raycast all objects each time. To speed up, we raycast all objects only once to find visible objects, and avoid raycasting to invisible objects in the remaining steps on this frame. As in~\cite{McCormac:etal:3DV2018}, we only raycast voxels whose foreground probability is higher than 0.5. Objects whose voxels are not raycasted at all are considered to be invisible from this view. 

 
\section{Experiments}
We evaluate our system on a Linux system with an Intel Core i7-7700 CPU at 3.50GHz with 32GB memory. Mask R-CNN segmentation is pre-computed on the GPU using the publicly available weights and implementation~\cite{Wu:etal:Tensorpack2016} without fine-tuning. Each object is stored in a separate octree-based volumetric model, modified based on source code of Supereight~\cite{Vespa:etal:RAL2018}.

\subsection{Robust camera pose estimation}
We first evaluate the camera tracking accuracy in dynamic environments using the widely used TUM RGB-D dataset~\cite{Sturm:etal:IROS2012}. The dataset provides RGB-D sequences with ground truth camera trajectory, recorded by a motion capture system. We report the commonly used Root-Mean-Square-Error (RMSE) of the Absolute Trajectory Error (ATE). To evaluate the effect of different cameras motion and environment change conditions, 6 different sequences are investigated. 
We compare our method with five state-of-the-art dynamic SLAM approaches: joint visual odometry and scene flow (VO-SF)~\cite{Jaimez:etal:ICRA2017}, StaticFusion (SF)~\cite{Scona:etal:ICRA2018}, DynaSLAM (DS)~\cite{Bescos:etal:RAL2018}, Co-Fusion (CF)~\cite{Runz::Agapito::ICRA2017}, and MaskFusion (MF)~\cite{Runz::Agapito::ARXIV2018}. VO-SF~\cite{Jaimez:etal:ICRA2017}, SF~\cite{Scona:etal:ICRA2018}, and DS~\cite{Bescos:etal:RAL2018} were designed for reconstructing the static background with dynamic parts ignored (or even inpainted as in DS~\cite{Bescos:etal:RAL2018}). CF~\cite{Runz::Agapito::ICRA2017} and MF~\cite{Runz::Agapito::ARXIV2018} were designed for multi-object reconstruction. In all these methods, DS~\cite{Bescos:etal:RAL2018} is the only method using feature-based sparse tracking (not reconstructing moving objects at all), while the remaining ones use dense tracking methods as ours. For fair comparison, we compare first with dense tracking methods and take DS~\cite{Bescos:etal:RAL2018} as an additional reference. Table~\ref{table: ate-compare} reports our experimental results.

\begin{figure}[tb]
	\centering
		\includegraphics[width=\onecolume]{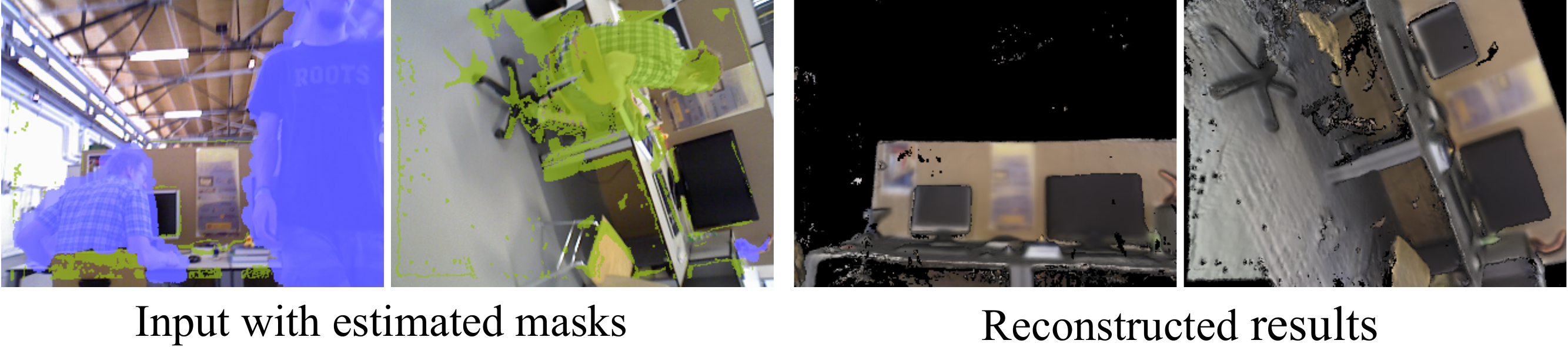}
		\vspace{-2mm}

   	\caption{Robust camera tracking and background reconstruction in a dynamic environment (in ``f3w halfsphere'' sequence). Moving persons are rejected due to the semantic labelling of Mask R-CNN (in blue) or during motion refinement (in green).}
	\vspace{-4mm}
	\label{fig:exp-tum}
\end{figure}

From the table~\ref{table: ate-compare}, we can see that our system achieves best results in almost all sequences among dense tracking method. Our method even outperforms VO-SF and SF, which were designed especially for robust camera tracking in a dynamic environment. Fig.~\ref{fig:exp-tum} shows two inputs and the reconstruction results in the challenging ``f3w halfsphere'' sequence. We highlight rejected segmentation masks in the input images, with the geometrically refined mask labelled as human in blue and the high residual regions during motion refinement in green. It can be noted in Fig.~\ref{fig:exp-tum} that even when Mask R-CNN fails to recognise a person, our combined segmentation using motion refinement can still reject it. DynaSLAM achieves best tracking accuracy in almost all sequences while it is the only sparse feature-based SLAM system among the tested approaches. It shows the potential advantage of feature trackers in dynamic environments. As part of future work, it would be very interesting to combine a feature-based approach with direct dense tracking/mapping methods to further improve camera tracking accuracy and robustness. This could also help to overcome current failure-cases in challenging conditions such as very reflective scenes or fast motions.

\begin{table}[tbp]
	\caption{\label{table: ate-compare} Quantitative comparison of camera tracking}
	\begin{tabular}{|l|c|c|c|c|c||c|c|}
		\hline
		\multicolumn{1}{|c|}{\multirow{2}{*}{Sequence}} & \multicolumn{6}{c|}{ATE RMSE (cm)}                 \\ \cline{2-7} 
		\multicolumn{1}{|c|}{}                          & VO-SF & SF  & CF   & MF   & \textbf{Ours} & DS* \\ \hline
		f3s static                                      & 2.9   & 1.3  & 1.1  & 2.1  & \textbf{1.0}  & \textbf{-}      \\
		f3s xyz                                         & 11.1  & 4.0  & \textbf{2.7} & 3.1  & 6.2  & \textbf{1.5}      \\
		f3s halfsphere                                  & 18.0  & 4.0  & 3.6  & 5.2  & \textbf{3.1}  & \textbf{1.7}       \\
		f3w static                                      & 32.7  & \textbf{1.4}  & 55.1 & 3.5  & 2.3 & \textbf{0.6}      \\
		f3w xyz                                         & 87.4  & 12.7 & 69.6 & 10.4 & \textbf{6.8} & \textbf{1.5}    \\
		f3w halfsphere                                  & 73.9  & 39.1  & 80.3 & 10.6 & \textbf{3.8} & \textbf{2.5}     \\ \hline
	\end{tabular}
	\vspace{1mm}

*: feature-based sparse approach. The others are dense-tracking approaches.
	\vspace{-3mm}

\end{table}

\subsection{Object reconstruction evaluation for other components}
We also tested our method within a fully controlled synthetic environment using photo-realistic rendering and trajectory simulation~\cite{Li:etal:BMVC2018}. We selected a typical indoor scene with a sofa and a chair being translated and rotated in front of the camera. We implicitly evaluate object pose estimation accuracy via object reconstruction error. 

\begin{figure}[tb]
	\centering
	\includegraphics[width=80mm]{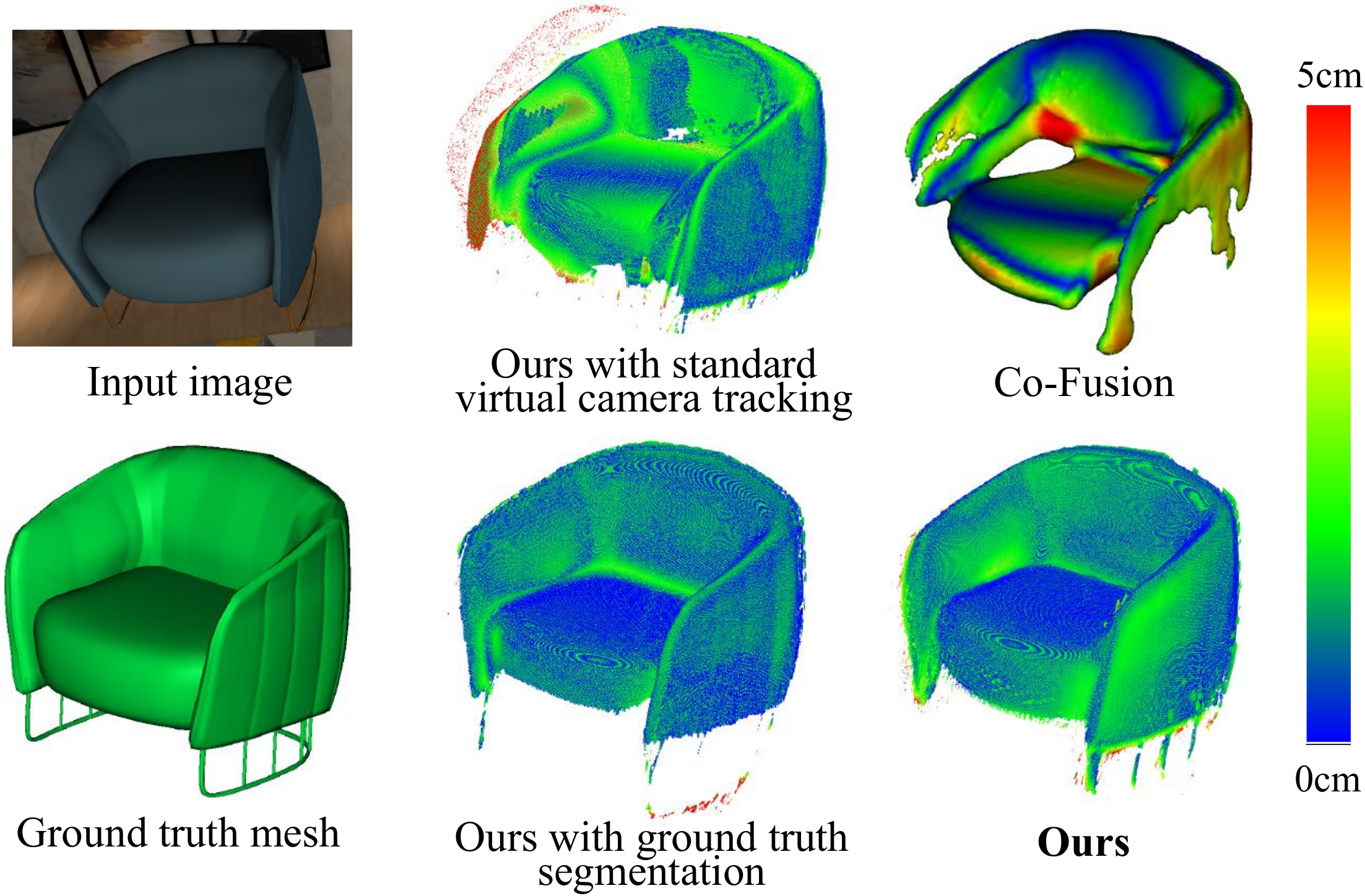}
		\vspace{-2mm}
	\caption{Comparison of reconstruction error for a moving sofa.}
	\label{fig:compare-reconstruction}
	\vspace{-5mm}
\end{figure}

To evaluate the effect of segmentation, we replaced the segmentation pipeline with ground truth masks (G.T.\ Seg.). We also compared our object-oriented tracker with virtual camera (V.C.) tracking to see if our parametrisation improves tracking accuracy. We further compare with Co-Fusion(CF)~\cite{Runz::Agapito::ICRA2017} using their public code. Table~\ref{table: ate-object-compare} reports the mean and standard deviation of reconstruction error in these experiments. The results show that our system can achieve more accurate object reconstructions. The difference between using ground truth masks and our own segmentation component is  negligible in the specific example. The higher error obtained using virtual camera tracking demonstrates the reliability of our object-centric tracking, especially for large object rotations. Fig. \ref{fig:compare-reconstruction} shows the visualisation comparison results on the sofa reconstruction.

\begin{table}[t]
\centering
	\caption{\label{table: ate-object-compare} Object reconstruction error (avg./std., in cm)}
	\vspace{-2mm}
	\begin{tabular}{c|c|c|c||c}
		\hline
		Method             & CF      & \begin{tabular}[c]{@{}c@{}}Ours with  \\ v.c. tracking\end{tabular}    & \textbf{Ours} & \begin{tabular}[c]{@{}c@{}}Ours with  \\ GT-Seg\end{tabular} \\ \hline
		Sofa &  1.72/1.62  & 1.68/1.90     & \textbf{0.74/0.79}  & 0.46/0.59 \\ \hline
		Chair    &  1.19/\textbf{1.33}  &  1.13/1.58                    &   \textbf{1.00}/1.66     &  0.92/1.74   \\ \hline
	\end{tabular}
\end{table}

\subsection{Real-world applications}
\begin{figure}[t]
	\centering
	\includegraphics[width=\linewidth]{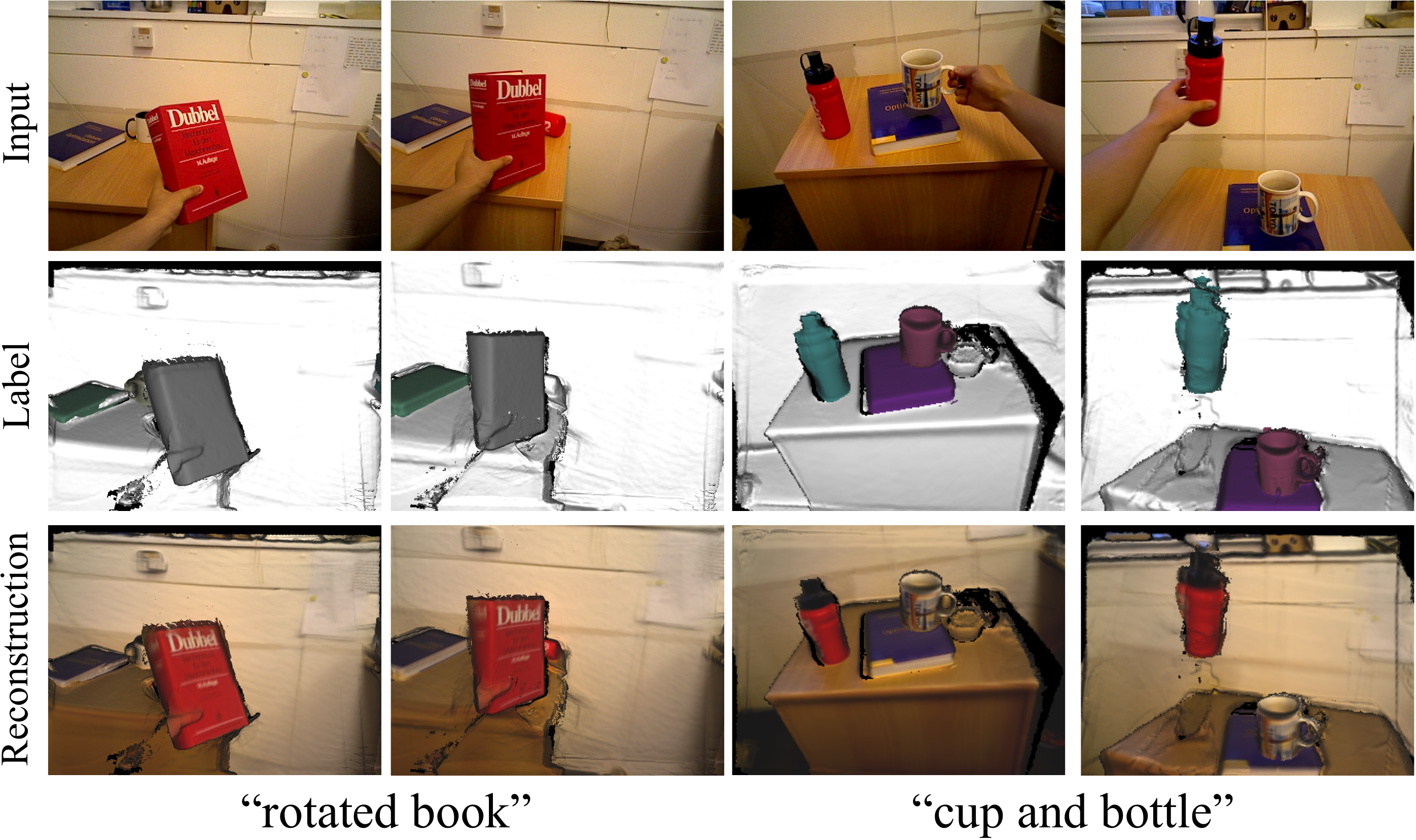}
	\vspace{-7mm}
	\caption{Qualitative demonstration: input RGB (top row), semantic class prediction (middle row) and geometry reconstruction result (bottom row).}
	\label{fig:demo}
\end{figure}

\begin{figure}[tb]
	\centering
	\includegraphics[width=\linewidth]{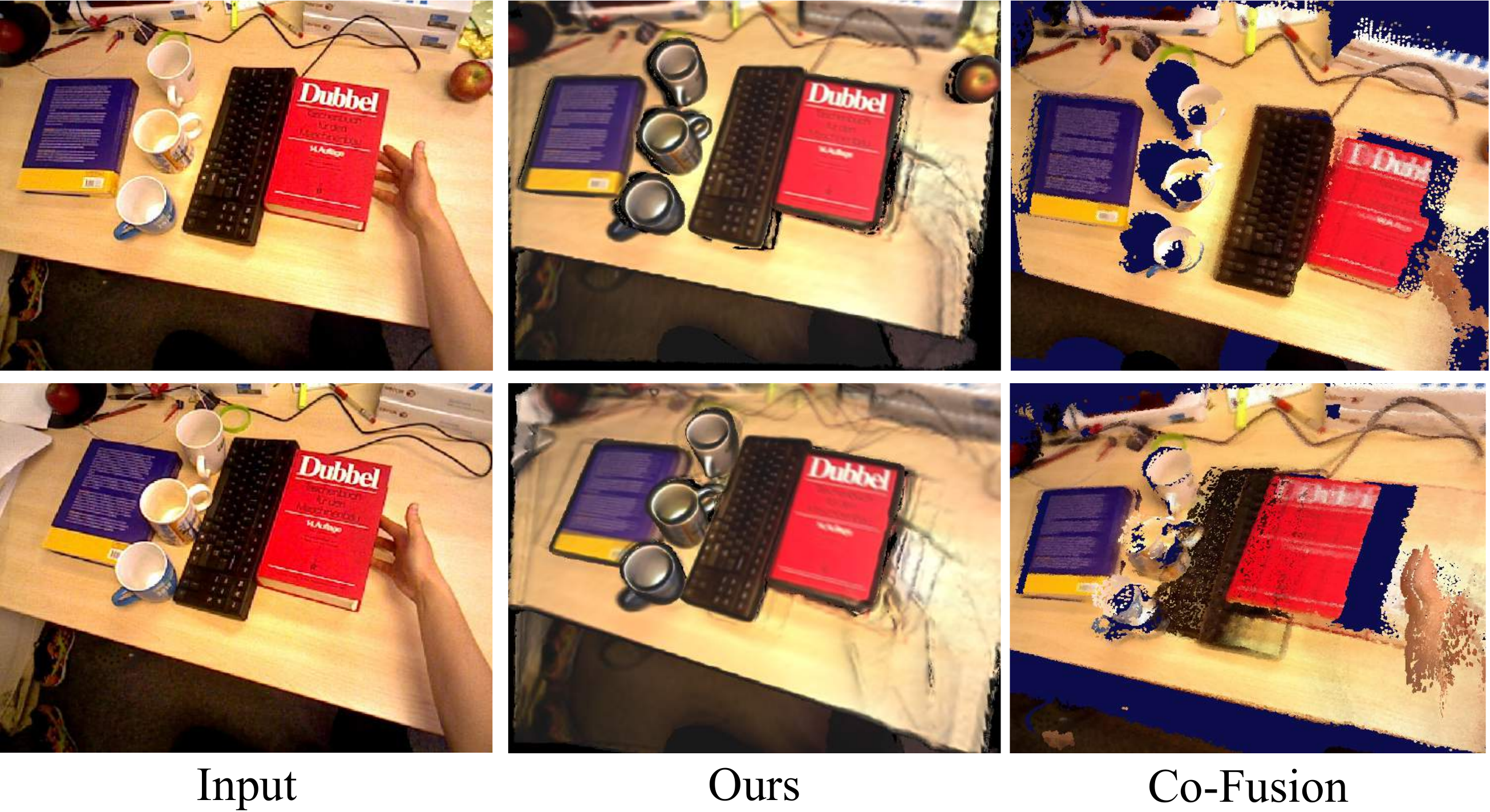}
	\vspace{-7mm}
	\caption{Qualitative comparison with Co-Fusion: input RGB (left column), our reconstruction results (middle) and Co-Fusion results (right column).}
	\label{fig:demo-multi}
	\vspace{-3mm}
\end{figure}

We demonstrated our proposed method in various scenarios to show its capabilities. Fig.~\ref{fig:demo} shows the results in two scenes， ``rotated book'' and ``cup and bottle''. For each input image, we provide label image and reconstruction to show the detailed reconstruction, reliable tracking, and segmentation. With separate volumetric maps for each object, our object models do not collide with each other, which is more suitable for multiple instance SLAM than a surfel-based system.
Fig.~\ref{fig:demo-multi} also shows a scene where our system can simultaneously support the robust tracking of more than 6 moving objects while maintaining a highly detailed reconstruction. As a qualitative comparison, we also show the reconstruction from Co-Fusion, which did not segment and reconstruct these moving objects successfully because the motion was not of sufficient magnitude. In addition, surfel-based systems, such as Co-Fusion and MaskFusion, do not provide the same level of details per object. On the contrary, our system can maintain highly detailed reconstructions and nevertheless keep efficient memory usage thanks to the octree data structure. More results can be seen in the video attachment. 

\subsection{Runtime analysis}

\begin{table}[tb]
	\vspace{-1mm}
	\centering
	\caption{Run-time analysis of system components (ms)}
	\vspace{-2mm}
	\label{table: time-system} 
	\begin{tabular}{c|c|c|c|c}
		\hline
		Components & Tracking & Segmentation & Integration & Raycasting  \\ \hline
		Time (ms)  &  43/MO     &   10/VO        &   12/VO.          &   8/VO         \\ \hline
	\end{tabular}
\end{table}

We evaluated the average computational time for the components of our dynamic SLAM system in different sequences with approximately 3 to 6 objects being moved. Processing time (all on CPU) for each frame averages 400 ms with more than 25 objects being generated in the scene. When a new object is detected, the initialisation takes around 10 ms per object. Tracking time scales mainly with moving objects (MO) while segmentation, integration and raycasting scales with visible objects (VO). A more-detailed breakdown of computation time for each component is shown in Table~\ref{table: time-system}.

We would like to highlight that our current system only runs on CPU without being highly optimised for performance yet. We believe a high frame-rate version of our system is achievable by exploiting GPU parallelisation. 

\section{Conclusions}
We present a novel approach for multi-instance dynamic SLAM using an octree-based volumetric representation. It robustly tracks camera pose in dynamic environment and continuously estimates dense geometry, semantics, and object foreground probabilities. Experimental results in various scenarios demonstrate the effectiveness of our method in indoor environments. We hope our method paves the way for new applications in indoor robotic applications, where an awareness of environment change, free space, and object-level information will empower the next generation of mobile robots.

\section*{ACKNOWLEDGMENTS}
We wish to thank John McCormac and Emanuele Vespa for fruitful discussions. This research is supported by Imperial College London and  the EPSRC grant Aerial ABM EP/N018494/1. Binbin Xu holds a China Scholarship Council-Imperial Scholarship.

\bibliographystyle{IEEEtran}
\bibliography{robotvision}

\end{document}